\pdfoutput=1

\documentclass[11pt]{article}

\usepackage{naacl2021}

\usepackage{times}
\usepackage{latexsym}
\usepackage{multirow}
\usepackage{color}
\usepackage{graphicx} 

\usepackage[T1]{fontenc}

\usepackage[utf8]{inputenc}

\usepackage{microtype}
\usepackage{booktabs}
\usepackage{amsmath,amssymb}
    \DeclareMathOperator*{\argmax}{arg\,max}
\usepackage{graphics}
\usepackage{bm}

\title{End-to-end Biomedical Entity Linking \\with Span-based Dictionary Matching}

\author{Shogo Ujiie$^{\spadesuit}$ \quad Hayate Iso$^{\heartsuit}$\thanks{~~Work done while at Nara Institute of Science and Technology.} \\
\textbf{Shuntaro Yada$^{\spadesuit}$ \quad Shoko Wakamiya$^{\spadesuit}$ \quad Eiji Aramaki$^{\spadesuit}$}\\
$^\spadesuit$Nara Institute of Science and Technology  \quad
$^\heartsuit$Megagon Labs\\
\texttt{\{ujiie, yada-s, wakamiya, aramaki\}@is.naist.jp}\\
\texttt{hayate@magagon.ai}
}

\begin{document}
\maketitle
\begin{abstract}
Disease name recognition and normalization
, which is generally called biomedical entity linking, 
is a fundamental process in biomedical text mining.
Recently, neural joint learning of both tasks has been proposed to utilize the mutual benefits.
While this approach achieves high performance, disease concepts that do not appear in the training dataset cannot be accurately predicted.
This study introduces a novel end-to-end approach that combines span representations with dictionary-matching features to address this problem.
Our model handles unseen concepts by referring to a dictionary while maintaining the performance of neural network-based models, in an end-to-end fashion.
Experiments using two major datasets demonstrate that our model achieved competitive results with strong baselines, especially for unseen concepts during training.

\end{abstract}

\section{Introduction}

Identifying disease names
, which is generally called biomedical entity linking, 
is the fundamental process of biomedical natural language processing, and it can be utilized in applications such as a literature search system ~\cite{Lee2016-fu} and a biomedical relation extraction ~\cite{Xu2016-ha}.
The usual system to identify disease names consists of two modules: named entity recognition (NER) and named entity normalization (NEN).
NER is the task that recognizes the span of a disease name, from the start position to the end position.
NEN is the post-processing of NER, normalizing a disease name into a controlled vocabulary, such as a MeSH or Online Mendelian Inheritance in Man (OMIM).

Although most previous studies have developed pipeline systems, in which the NER model first recognizs disease mentions~\cite{Lee2020-ne, Weber2020-zu} and the NEN model normalizes the recognized mention~\cite{Leaman2013-lm, Ferre2020-wl, Xu2020-nj, medtype2020}, a few approaches employ a joint learning architecture for these tasks~\cite{Leaman2016-ji,Lou2017-fd}.
These joint approaches simultaneously recognize and normalize disease names utilizing their mutual benefits.
For example, ~\citet{Leaman2013-lm} demonstrated that dictionary-matching features, which are commonly used for NEN, are also effective for NER.
While these joint learning models achieve high performance for both NER and NEN,
they predominately rely on hand-crafted features, 
which are difficult to construct because of the domain knowledge requirement.

Recently, a neural network (NN)-based model that does not require any hand-crafted features was applied to the joint learning of NER and NEN~\cite{Zhao_Liu_Zhao_Wang_2019}.
NER and NEN were defined as two token-level classification tasks, i.e., their model classified each token into IOB2 tags and concepts, respectively.
Although their model achieved the state-of-the-art performance for both NER and NEN, a concept that does not appear in training data (i.e., zero-shot situation) can not be predicted properly.

One possible approach to handle this zero-shot situation is utilizing the dictionary-matching features.
Suppose that an input sentence ``Classic \textit{polyarteritis nodosa} is a systemic vasculitis'' is given, where ``\textit{polyarteritis nodosa}'' is the target entity.
Even if it does not appear in the training data, it can be recognized and normalized by referring to a controlled vocabulary that contains ``\textit{Polyarteritis Nodosa} (MeSH: D010488).''
Combining such looking-up mechanisms with NN-based models, however, is not a trivial task; dictionary matching must be performed at the \textit{entity}-level, whereas standard NN-based NER and NEN tasks are performed at the \textit{token}-level \cite[for example, ][]{Zhao_Liu_Zhao_Wang_2019}.

\begin{figure*}[t]
  \begin{center}
    \includegraphics[clip,width=15.0cm]{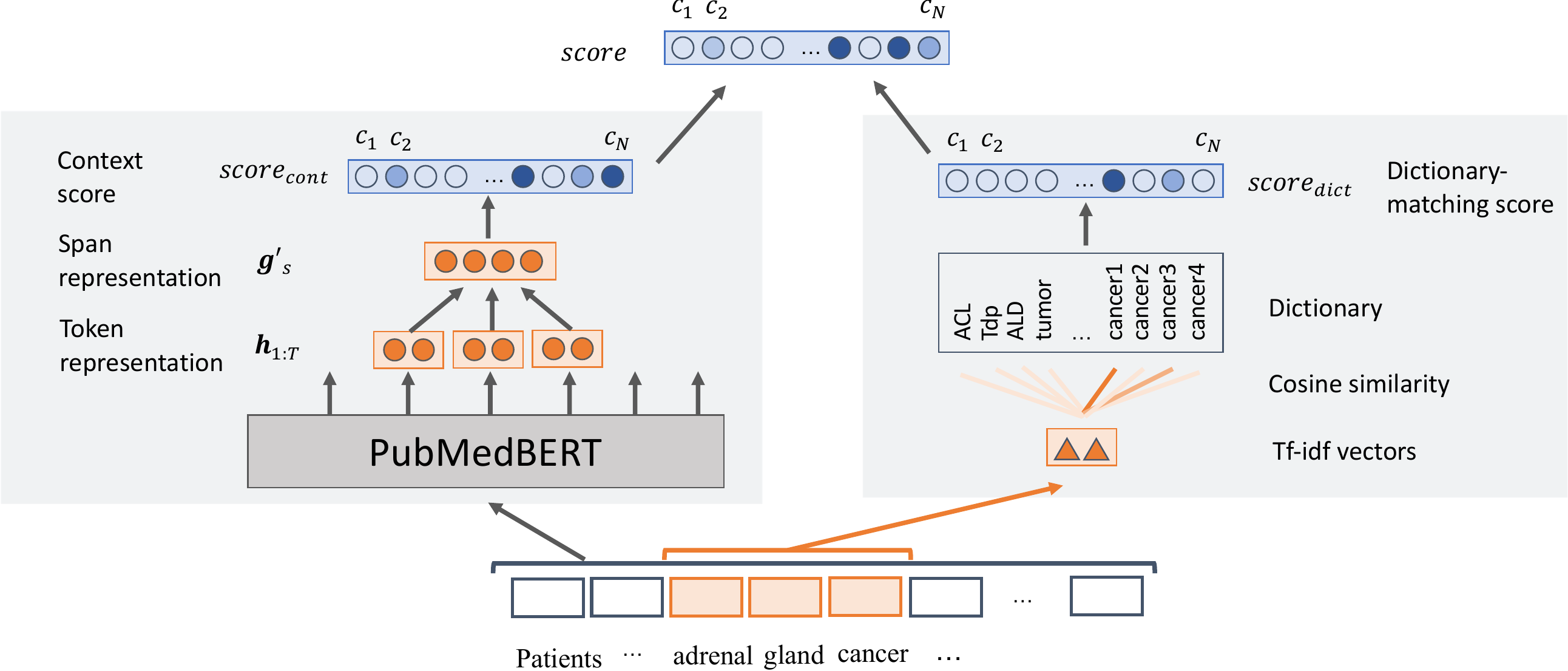}
    \caption{The overview of our model. It combines the dictionary-matching scores with the context score obtained from PubMedBERT. The red boxes are the target span and ``ci'' in the figure is the ``i''-th concept in the dictionary.}
    \label{fig:overview}
  \end{center}
\end{figure*}

To overcome this problem, we propose a novel end-to-end approach for NER and NEN that combines dictionary-matching features with NN-based models.
Based on the span-based model introduced by \citet{Lee2017-zp}, our model first computes span representations for all possible spans of the input sentence and then combines the dictionary-matching features with the span representations.
Using the score obtained from both features, it directly classifies the disease concept.
Thus, our model can handle the zero-shot problem by using dictionary-matching features while maintaining the performance of the NN-based models.

Our model is also effective in situations other than the zero-shot condition.
Consider the following input sentence:
    ``We report the case of a patient who developed acute \textit{hepatitis},''
where ``\textit{hepatitis}'' is the target entity that should be normalized to ``drug-induced hepatitis.''
While the longer span ``acute hepatitis'' also appears plausible for stand\-alone NER models, our end-to-end architecture assigns a higher score to the correct shorter span ``hepatitis'' due to the existence of the normalized term (``drug-induced hepatitis'') in the dictionary.

Through the experiments using two major NER and NEN corpora, we demonstrate that our model achieves competitive results for both corpora.
Further analysis illustrates that the dictionary-matching features improve the performance of NEN in the zero-shot and other situations.

Our main contributions are twofold:
({\romannumeral 1}) We propose a novel end-to-end model for disease name recognition and normalization that utilizes both NN-based features and dictionary-matching features;
({\romannumeral 2}) We demonstrate that combining dictionary-matching features with an NN-based model is highly effective for normalization, especially in the zero-shot situations.

\section{Methods}
\subsection{Task Definition}
Given an input sentence, which is a sequence of words $\boldsymbol{x} = \{x_1, x_2, \cdots, 
x_{|{\bm{X}}|}\}$ in the biomedical literature,
let us define $\mathcal{S}$ as a set of all possible spans, and $\mathcal{L}$ as a set of concepts that contains the special label $\mathit{Null}$ for a non-disease span.
Our goal
is to predict a set of labeled spans $\boldsymbol{y} = \{\langle i, j, d \rangle _k\}_{k=1}^{|{\bm{Y}}|}$, where $(i, j) \in \mathcal{S}$ is the word index in the sentence, and $d \in \mathcal{L}$ is the concept of diseases.

\subsection{Model Architecture}
Our model predicts the concepts for each span based on the score, which is represented by the weighted sum of two factors: the context score $score_{cont}$ obtained from span representations and the dictionary-matching score $score_{dict}$.
Figure \ref{fig:overview} illustrates the overall architecture of our model.
We denote the score of the span $s$ as follows:
\begin{align*}
    score (s, c) = score_{cont} (s, c) + 
    \lambda score_{dict} (s, c)
\end{align*}
where $c \in \mathcal{L}$ is the candidate concept and $\lambda$ is the hyperparameter that balances the scores.
For the concept prediction, the scores of all possible spans and concepts are calculated, and then the concept with the highest score is selected as the predicted concept for each span as follows:
\begin{equation*}
    y = \argmax_{c \in \mathcal{L}} score(s, c)
\end{equation*}

\paragraph{Context score}
The context score is computed in a similar way to that of ~\citet{Lee2017-zp}, which is based on the span representations.
To compute the representations of each span, the input tokens are first encoded into the token embeddings.
We used BioBERT~\cite{Lee2020-ne} as the encoder, which is a variation of bidirectional encoder representations from transformers (BERT) that is trained on a large amount of biomedical text.
Given an input sentence containing $T$ words, we can obtain the contextualized embeddings of each token using BioBERT as follows:
\begin{eqnarray*}
    {\mathbf{h}}_{1:T} = {\rm BERT} (x_1, x_2, \cdots, x_T)
\end{eqnarray*} 
where ${\mathbf{h}}_{1:T}$ is the input tokens embeddings.

Span representations are obtained by concatenating several features from the token embeddings:
\begin{eqnarray*}
    {\mathbf{g}}_s = [{\mathbf{h}}_{start(s)}, {\mathbf{h}}_{end(s)}, \hat{{\mathbf{h}}}_s, \phi(s)] \\
    {\mathbf{g'}}_s = {\rm GELU}({\rm FFNN}({\mathbf{g}}_s))
\end{eqnarray*}
where ${\mathbf{h}}_{start(s)}$ and ${\mathbf{h}}_{end(s)}$ are the start and end token embeddings of the span, respectively; and $\hat{{\mathbf{h}}_s}$ is the weighted sum of the token embeddings in the span, which is obtained using an attention mechanism~\cite{Bahdanau2015-mx}.
$\phi(i)$ is the size of span $s$.
These representations ${\mathbf{g}}_s$ are then fed into a simple feed-forward NN, ${\rm FFNN}$, and a nonlinear function, ${\rm GELU}$~\cite{Hendrycks2016-xl}.

Given a particular span representation and a candidate concept as the inputs, we formulate the context score as follows:
\begin{eqnarray*}
    score_{cont}(s, c) = {\mathbf{g}}_s \cdot {\mathbf{W}}_c
\end{eqnarray*}
where $\mathbf{W} \in \mathbb{R}^{|\mathcal{L}| \times d^{{\mathbf{g}}}}$ is the weight matrix associated with each concept $c$, and ${\mathbf{W}}_c$ represents the weight vector for the concept $c$.

\paragraph{Dictionary-matching score}
We used the cosine similarity of the TF-IDF vectors as the dictionary-matching features.
Because there are several synonyms for a concept, we calculated the cosine similarity for all synonyms of the concept and used the maximum cosine similarity as the score for each concept.
The TF-IDF is calculated using the character-level n-gram statistics computed for all diseases appearing in the training dataset and controlled vocabulary.
For example, given the span ``breast cancer,'' synonyms with high cosine similarity are ``breast cancer (1.0)'' and ``male breast cancer (0.829).''

\section{Experiment}
\subsection{Datasets}
To evaluate our model, we chose two major datasets used in disease name recognition and normalization against a popular controlled vocabulary, MEDIC \cite{Davis2012-pp}.
Both datasets, the National Center for Biotechnology Information Disease (NCBID) corpus \cite{Dogan2014-nv} and the BioCreative V Chemical Disease Relation (BC5CDR) task corpus  \cite{Li2016-gz}, comprise of PubMed titles and abstracts annotated with disease names and their corresponding normalized term IDs (CUIs).
NCBID provides 593 training, 100 development, and 100 test data splits, while BC5CDR evenly divides 1500 data into the three sets.
We adopted the same version of MEDIC as TaggerOne \cite{Leaman2016-ji} used, and that we dismissed non-disease entity annotations contained in BC5CDR.

\subsection{Baseline Models}
We compared several baselines to evaluate our model.
DNorm~\cite{Leaman2013-lm} and NormCo~\cite{noauthor_undated-bl} were used as pipeline models due to their high performance.
In addition, we used the pipeline systems consisting of state-of-the-art models: BioBERT~\cite{Lee2020-ne} for NER and BioSyn ~\cite{Sung2020-ra} for NEN.

TaggerOne~\cite{Leaman2016-ji} and Transition-based model~\cite{Lou2017-fd} are used as joint-learning models.
These models outperformed the pipeline models in NCBID and BC5CDR.
For the model introduced by \citet{Zhao_Liu_Zhao_Wang_2019}, we cannot reproduce the performance reported by them.
Instead, we report the performance of the simple token-level joint learning model based on the BioBERT, which referred as ``joint (token)''.

\subsection{Implementation}
We performed several preprocessing steps: splitting the text into sentences using the NLTK toolkit ~\cite{bird2009natural}, removing punctuations, and resolving abbreviations using Ab3P~\cite{Sohn2008-bj}, a common abbreviation resolution module.
We also merged disease names in each training set into a controlled vocabulary, following the methods of  ~\citet{Lou2017-fd}.

For training, we set the learning rate to 5e-5, and mini-batch size to 32.
$\lambda$ was set to 0.9 using the development sets.
For BC5CDR, we trained the model using both the training and development sets following ~\citet{Leaman2016-ji}.
For computational efficiency, we only consider spans with up to 10 words.

\begin{table}
\centering
\footnotesize
\begin{tabular}{lcccc}
    \toprule & \multicolumn{2}{c}{NCBID} & \multicolumn{2}{c}{BC5CDR} \\
    \cmidrule(r){2-3} \cmidrule(r){4-5}
    Models & NER & NEN & NER & NEN \\
    \midrule
    TaggerOne & 0.829 & 0.807 & 0.826 & 0.837\\
    Transition-based model & 0.821 & 0.826 & 0.862 & \textbf{0.876}\\
    NormCo & 0.829 & 0.840 & 0.826 & 0.830\\
    pipeline & 0.874 & 0.841 & 0.865 &  0.818\\
    joint (token) & 0.864 & 0.765 & 0.855 & 0.817\\
    \midrule
    Ours without dictionary & 0.884 & 0.781 & 0.864 & 0.808\\
    Ours & \textbf{0.891} & \textbf{0.854} & \textbf{0.867} & 0.851\\
    \bottomrule
\end{tabular}
\caption{\label{ONLY_F1} F1 scores of NER and NEN in NCBID and BC5CDR. Bold font represents the highest score.}
\end{table}

\subsection{Evaluation Metrics}
We evaluated the recognition performance of our model using micro-F1 at the entity level.
We consider the predicted spans as true positive when their spans are identical.
Following the previous work~\cite{noauthor_undated-bl, Leaman2016-ji}, the performance of NEN was evaluated using micro-F1 at the abstract level.
If a predicted concept was found within the gold standard concepts in the abstract, regardless of its location, it was considered as a true positive.

\section{Results \& Discussions}
Table \ref{ONLY_F1} illustrates that
our model mostly achieved the highest F1-scores in both NER and NEN, except for the NEN in BC5CDR, in which the transition-based model displays its strength as a baseline.
The proposed model outperformed the pipeline model of the state-of-the-art models for both tasks, which demonstrates that the improvement is attributed not to the strength of BioBERT but the model architecture, including the end-to-end approach and combinations of dictionary-matching features.

Comparing the model variation results, adding dictionary-matching features improved the performance in NEN.
The results clearly suggest that dictionary-matching features are effective for NN-based NEN models.

\begin{table}
\centering
\footnotesize
\begin{tabular}{lcccc}
    \toprule & \multicolumn{2}{c}{standard} & \multicolumn{2}{c}{zero-shot} \\
    \cmidrule(r){2-3} \cmidrule(r){4-5}
    dataset & mention & concept & mention & concept \\
    \midrule
    NCBID & 781 & 135 & 179 & 56\\
    BC5CDR & 4031 & 461 & 391 & 179\\
    \bottomrule
\end{tabular}
\caption{\label{STATISTICS} Number of mentions and concepts in standard and zero-shot situations.}
\end{table}

\begin{table}
    \centering
    \footnotesize
    \begin{tabular}{clcc}
    \toprule
    & Methods & NCBID & BC5CDR\\
    \midrule
    \multirow{2}{*}{zero-shot} & Ours without dictionary &  0 & 0 \\
        & Ours  & \textbf{0.704} & \textbf{0.597} \\
    \midrule
    \multirow{2}{*}{standard} & Ours without dictionary &  0.854 & 0.846\\
        & Ours  & \textbf{0.905} & \textbf{0.877}\\
    \bottomrule
    \end{tabular}
    \caption{F1 scores for NEN of NCBID and BC5CDR subsets for zero-shot situation where disease concepts do not appear in training data and the standard situation where they do appear in training data.}
    \label{tab:zero_norm}
\end{table}

\subsection{Contribution of Dictionary-Matching}

To analyze the behavior of our model in the zero-shot situation, we investigated the NEN performance on two subsets of both corpora: disease names with concepts that appear in the training data (i.e., standard situation), and disease names with concepts that do not appear in the training data (i.e., the zero-shot situation).
Table \ref{STATISTICS} shows the number of mentions and concepts in each situation.
Table \ref{tab:zero_norm} displays the results of the zero-shot and standard situation.
The proposed model with dictionary-matching features can classify disease concepts in the zero-shot situation, whereas the NN-based classification model cannot normalize the disease names.

The results of the standard situation demonstrate that combining dictionary-matching features also improves the performance even when target concepts appear in the training data.
This finding implies that an NN-based model can benefit from dictionary-matching features, even if the models can learn from many training data.

\subsection{Case study}
We examined 100 randomly sampled sentences to determine the contributions of dictionary-matching features.
There are 32 samples in which the models predicted concepts correctly by adding dictionary-matching features.
Most of these samples are disease concepts that do not appear in the training set but appear in the dictionary.
For example, ``\textit{pure red cell aplasis} (MeSH: D012010)” is not in the BC5CDR training set while the MEDIC contains ``Pure Red-Cell Aplasias” for “D012010”.
In this case, a high dictionary-matching score clearly leads to a correct prediction in the zero-shot situation.

In contrast, 
there are 32 samples in which the dictionary-matching features cause errors.
The sources of this error type are typically general disease names in the MEDIC.\@
For example, ``Death (MeSH:D003643)'' is incorrectly predicted as a disease concept in NER.
Because these words are also used in the general context, %
our model overestimated their dictionary-matching scores. 

Furthermore, in the remaining samples, our model predicted the code properly and the span incorrectly.
For example, although ``thoracic hematomyelia'' is labeled as ``MeSH: D020758'' in the BC5CDR test set, our model recognized this as ``hematomyelia.''
In this case, our model mostly relied on the dictionary-matching features and mis-classifies the span because `hematomyelia'' is in the MEDIC but not in the training data.

\subsection{Limitations}
Our model is inferior to the transition-based model for BC5CDR.
One possible reason is that the transition-based model utilizes normalized terms that co-occur within a sentence, whereas our model does not.
Certain disease names that co-occur within a sentence are strongly useful for normalizing disease names.
Although BERT implicitly considers the interaction between disease names via the attention mechanism, a more explicit method is preferable for normalizing diseases.

Another limitation is that our model treats the dictionary entries equally.
Because certain terms in the dictionary may also be used for non-disease concepts, such as gene names, we must consider the relative importance of each concept.

\section{Conclusion}
We proposed a end-to-end model for disease name recognition and normalization that combines the NN-based model with the dictionary-matching features.
Our model achieved highly competitive results for the NCBI disease corpus and BC5CDR corpus, demonstrating that incorporating dictionary-matching features into an NN-based model can improve its performance.
Further experiments exhibited that dictionary-matching features enable our model to accurately predict the concepts in the zero-shot situation, and they are also beneficial in the other situation.
While the results illustrate the effectiveness of our model, we found several areas for improvement, such as the general terms in the dictionary and the interaction between disease names within a sentence.
A possible future direction to deal with general terms is to jointly train the parameters representing the importance of each synonyms.

\bibliography{naacl2021}
\bibliographystyle{acl_natbib}

\end{document}